\documentclass[letterpaper, 10 pt, conference]{ieeeconf}  

\IEEEoverridecommandlockouts                              

\usepackage{hyperref}
\usepackage{float}
\usepackage{verbatim} 
\usepackage{array}
\restylefloat{figure}
\restylefloat{table}
\usepackage{multirow}
\usepackage{amsmath}
\usepackage{amssymb}
\usepackage{booktabs}
\usepackage{graphics}
\usepackage{multirow}
\usepackage{hhline}
\usepackage{makecell}
\usepackage{siunitx}
\usepackage{array}
\usepackage{vcell}
\usepackage{float}
\usepackage{adjustbox}
\usepackage{tabularray}
\usepackage{subcaption}
\usepackage{xcolor}
\usepackage{tabularx}
\usepackage{graphicx}
\usepackage{pifont}
\usepackage{amssymb}
\newcommand{\cmark}{\ding{51}}%
\newcommand{\xmark}{\ding{55}}%
\usepackage[flushleft]{threeparttable}

\title{\LARGE \bf
3DEG: Data-Driven Descriptor Extraction for Global re-localization in subterranean environments}

\author{Nikolaos Stathoulopoulos$^1$, Anton Koval$^1$ and George Nikolakopoulos$^1$ 
\thanks{This work has been partially funded by the European Unions Horizon
2020 Research and Innovation Programme under the Grant Agreements No.
869379 illuMINEation, No.101003591 NEXGEN-SIMS.}
\thanks{$^{1}$The Authors are with the Robotics and AI Group, Department of Computer, Electrical and Space Engineering, Lule\r{a} University of Technology, 971 87 Lule\r{a}, Sweden} %
\thanks{Corresponding Author's Email: \texttt{niksta@ltu.se}} %
}

\begin{document}

\maketitle
\thispagestyle{empty}
\pagestyle{empty}

\begin{abstract}

Localization algorithms that rely on 3D LiDAR scanners often encounter temporary failures due to various factors, such as sensor faults, dust particles, or drifting. These failures can result in a misalignment between the robot’s estimated pose and its actual position in the global map. To address this issue, the process of global re-localization becomes essential, as it involves accurately estimating the robot’s current pose within the given map. In this article, we propose a novel global re-localization framework that addresses the limitations of current algorithms heavily reliant on scan matching and direct point cloud feature extraction. Unlike most methods, our framework eliminates the need for an initial guess and provides multiple top-$k$ candidates for selection, enhancing robustness and flexibility. Furthermore, we introduce an event-based re-localization trigger module, enabling autonomous robotic missions. Focusing on subterranean environments with low features, we leverage range image descriptors derived from 3D LiDAR scans to preserve depth information. Our approach enhances a state-of-the-art data-driven descriptor extraction framework for place recognition and orientation regression by incorporating a junction detection module that utilizes the descriptors for classification purposes. 
The effectiveness of the proposed approach was evaluated across three distinct real-life subterranean environments.

\end{abstract}


\section{Introduction} \label{introduction}
\begin{figure*}[h!] 
    \includegraphics[width=0.32\textwidth]{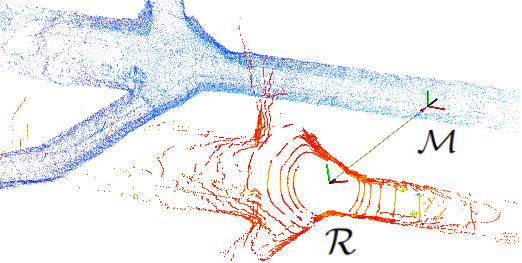}
    \hfill
    \includegraphics[width=0.32\textwidth]{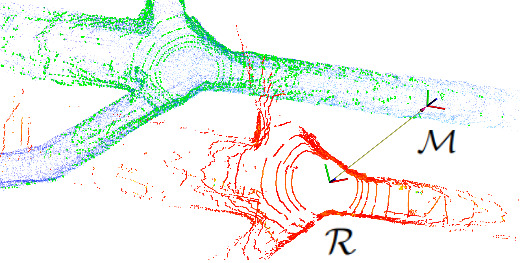}
    \hfill
    \includegraphics[width=0.32\textwidth]{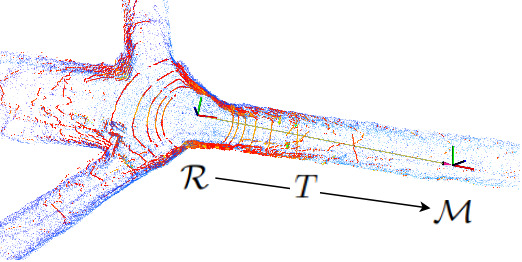}
    \caption{(a) Initially, the LiDAR scan of the robot frame $\mathcal{R}$ and the point cloud map of the map frame $\mathcal{M}$ are not aligned. (b) During the re-localization process, the nearest submap based on the indexes of the vectors in the $k$-d tree, highlighted in green, is selected to find the transform $T$. (c) Finally, the LiDAR scan of the robot frame is aligned with the map frame.}
    \label{fig:matching}
\end{figure*}
In recent years, there has been a growing emphasis on the exploration of GPS-denied environments using autonomous robots. These environments pose unique challenges, including harsh conditions, poor illumination, lack of structure, and uncharted territories. Consequently, there is an increasing demand for robust algorithms that can effectively navigate and explore these challenging environments while ensuring the safety of human operators~\cite{nikolakopoulos2021pushing, LINDQVIST2022104134}.
In the context of exploration or navigation missions, having access to a reliable global map is crucial. Such a map provides valuable information for tasks such as path planning, coordination of multiple robots, and localization of objects and survivors in Search And Rescue (SAR) missions. However, even with a global map, localization algorithms often encounter temporary failures due to various factors, including sensor faults, dust particles, or drifting. These failures can result in a misalignment between the robot's estimated pose and its actual position in the global map. To address this issue, the process of global re-localization becomes essential, as it involves accurately estimating the robot's current pose within the given map. This enables the resumption of missions in previously mapped environments or the correction of misalignment issues, ensuring the reliability and accuracy of the robot's navigation.
While traditional approaches for place recognition heavily rely on camera images due to their rich and descriptive information, they often struggle with environment changes and are not well-suited for low-light applications~\cite{kominiak2020mav, agha2021nebula}. In contrast, LiDAR sensors offer distinct advantages in challenging environments. They are immune to appearance changes and variations in illumination, making them highly reliable and robust sensing devices~\cite{LIO-SAM}. Moreover, recent advancements in deep learning techniques have facilitated the development of efficient data representations and feature descriptors for LiDAR point clouds. These advancements have greatly improved the performance of LiDAR-based methods in computer vision tasks, including global re-localization~\cite{OREOS, DH3D}. By leveraging these LiDAR-based approaches, the limitations associated with environment changes and low-light conditions can be overcome, making them a viable and promising choice for addressing the challenges of global re-localization and other computer vision tasks in complex and unexplored environments.


\subsection{Related work} \label{sec:related}

Predominantly, the global re-localization problem consists of two stages: (a) place recognition, which identifies the frame in the map that is topologically close to the current frame, and (b) pose estimation, which calculates the relative pose from the map frame to the robot's current frame. Our proposed framework serves as a bridge between descriptor extraction for place recognition and global pose estimation within a prebuilt point cloud map. Therefore, we structure this section into discussions of related work concerning learned descriptors for place recognition and pose estimation, as well as available global re-localization solutions in a point cloud map.

\subsubsection{Learned Descriptors}

Descriptors can be categorized as handcrafted \cite{M2DP, ScanContext}, learned-based \cite{OREOS, DH3D, PointNetVLAD, LoGG3D-Net, SegMap}, or hybrid \cite{Locus}. Handcrafted methods have the advantage of not requiring re-training to adapt to different environments or platforms. While methods like the well-known ScanContext by~\cite{ScanContext} have demonstrated reliable performance across varying scenarios, the discriminatory ability of such methods remains limited.
More recently, in the context of urban autonomous driving,~\cite{contour_context} introduced Contour Context, a novel approach for topological loop closure detection and accurate 3-DoF metric pose estimation. The proposed method leverages the layered distribution of structures within Cartesian birds' eye view (BEV) images, obtained from 3D LiDAR points. By extracting contour information from these images and assessing their geometric consistency and similarity, the approach achieves effective place recognition while optimizing relative transforms.

Learned-based methods have shown promising results with the universal approximation function properties of neural networks~\cite{nn_approx}. In recent years, CNNs have become the state-of-the-art method for generating learning-based descriptors due to their ability to find complex patterns in data~\cite{ImageNet}. PointNetVLAD, proposed by~\cite{PointNetVLAD}, pioneered the use of an end-to-end trainable global descriptor for 3D point cloud recognition. Extracted local features from PointNet~\cite{PointNet} are deployed to the NetVLAD aggregator~\cite{NetVLAD} to form a global descriptor of the scene. SegMap~\cite{SegMap} employs CNNs to encode small-dimensional representations and decode them back to the original input, as part of its core modules: segment extraction, description, localization, map reconstruction, and semantic extraction, all contributing to 3D point cloud localization and mapping.
LoGG3D-Net~\cite{LoGG3D-Net}, for the first time, addressed the limitations of first-order aggregation by introducing a training signal to the local features and using differentiable second-order pooling for global descriptor generation. High-order aggregation methods demonstrated superior performance in visual recognition~\cite{second_order, higher_order}, previously applied to 3D place recognition~\cite{Locus}, though not in a trainable architecture.
In DH3D~\cite{DH3D}, a hierarchical 3D descriptor learning approach was presented. A hierarchical network, operating directly on a point cloud, delivers local descriptors, a keypoint score map, and a global descriptor in a single forward pass. The success of deep learning is particularly noticeable in 2D images, where convolutional kernels can be easily applied to the 2D grid structure of the image. \cite{OREOS}, with OREOS, takes advantage of this success by projecting a 3D point cloud into spherical coordinates, yielding a 360-degree range image. The learned data-driven descriptors are then used for fetching the nearest neighbor place and estimating yaw discrepancy.
Similarly to the preceding approaches, yet harnessing a diverse array of cues encompassing range, normals, intensity, and semantic classes, the OverlapNet framework by \cite{chen2020overlapnet} effectively exploits spherical images derived from point cloud data, resulting in a notable enhancement of its performance. Expanding upon this foundational work, the advanced OverlapTransformer~\cite{ma2022overlaptrans} iteration introduces the integration of rotation-invariant features and expedited inference capabilities. Achieving this is facilitated through the incorporation of the attention mechanism derived from the Transformer~\cite{attention} and the NetVLAD head~\cite{NetVLAD}.

Finally, as the name suggests, hybrid methods aim to unite mathematical models with data-driven models to benefit from both~\cite{model_based_deep_learning}. First demonstrated by Locus~\cite{Locus}, an approach for LiDAR-based place recognition mathematically models topological relationships and temporal consistencies of point segments, while structural visual aspects of the segments were encoded using a data-driven 3D-CNN. Although it achieved state-of-the-art performance on the KITTI dataset~\cite{KITTI}, it struggles to adapt to environments where the extracted segments are structurally different from its training data. With LocNet~\cite{LocNet}, ~\cite{LocNet_Yin} used semi-handcrafted range histogram features as input to a 2D Convolutional Neural Network (CNN), demonstrating the power of Deep Neural Networks (DNNs) to learn suitable data representations and exploit the most relevant cues in the input data.

\subsubsection{Global re-localization}

Currently, only a limited number of ROS packages support global re-localization in a 3D point cloud map. \cite{hdl_localization} has provided a series of packages that include global re-localization as part of the localization and mapping process. The localization process employs an Unscented Kalman Filter-based pose estimation, fusing IMU and 3D LiDAR data.
Subsequently, the scheme performs Normal Distribution Transform (NDT) scan matching between the global map and the input scan to correct the estimated pose. For global re-localization, it offers three engines: Branch and Bounce Search (BBS)\cite{BBS}, FPFH+RANSAC\cite{FPFH, FPFH_2}, or FPFH+Teaser++~\cite{Teaser++}.

In the case of LIO-SAM~\cite{LIO-SAM}, re-localization based on LIO-SAM employs multiple factors, including IMU data, 3D LiDAR data, and the loop closure process, to jointly optimize the factor graph. This approach introduces key-frames and a sliding window scan-matching strategy, where new key-frames are selectively registered to a fixed-size set of prior sub-key-frames to enhance real-time performance.
The most recent package developed is based on FAST-LIO~\cite{FAST-LIO}. It provides global pose estimation in a pre-built point cloud map by combining low-frequency global localization and high-frequency odometry. The feature extraction process involves extracting edge and planar features from the input 3D LiDAR scan, which, along with IMU measurements, are fed into the state estimation module.
\begin{table*}[!t]
\caption{Comparison of the proposed framework against other available place-recognition solutions and re-localization frameworks. \textit{(NA: Not Applicable)
}} \label{table:comparison}
\centering
\resizebox{\linewidth}{!}{%
\begin{tblr}{
  row{1} = {c},
  cell{2}{2} = {c},
  cell{2}{3} = {c},
  cell{2}{4} = {c},
  cell{2}{5} = {c},
  cell{2}{6} = {c},
  cell{2}{7} = {c},
  cell{3}{2} = {c},
  cell{3}{3} = {c},
  cell{3}{4} = {c},
  cell{3}{5} = {c},
  cell{3}{6} = {c},
  cell{3}{7} = {c},
  cell{4}{2} = {c},
  cell{4}{3} = {c},
  cell{4}{4} = {c},
  cell{4}{5} = {c},
  cell{4}{6} = {c},
  cell{4}{7} = {c},
  cell{5}{2} = {c},
  cell{5}{3} = {c},
  cell{5}{4} = {c},
  cell{5}{5} = {c},
  cell{5}{6} = {c},
  cell{5}{7} = {c},
  cell{6}{2} = {c},
  cell{6}{3} = {c},
  cell{6}{4} = {c},
  cell{6}{5} = {c},
  cell{6}{6} = {c},
  cell{6}{7} = {c},
  hline{1-2,7} = {-}{0.08em},
}
                 & OREOS & PointNetVLAD & 3DEG (OURS) & LIO-SAM & FAST-LIO & HDL Localization \\
Input format & Data sequences & Data sequences & PCD file & PCD file & PCD file & PCD file \\
Initial guess & Not required & Not required & Not required & Required & Required & Required  \\
Yaw regression   & \cmark & \xmark & \cmark & NA & NA & NA        \\
Top-k candidates & \cmark & \cmark & \cmark & \xmark & \xmark      &    \xmark         \\
Event-triggered  & NA & NA & \cmark & \xmark & \xmark & \xmark        
\end{tblr}
}
\end{table*}

\subsection{Overview of the proposed approach}
Our proposed re-localization framework, referred to as 3DEG, acts as a pivotal link between existing place recognition methods and available integrated re-localization solutions, as illustrated in Table~\ref{table:comparison}. We address the limitations of conventional place recognition methods, which often depend on sequential data rather than a global point cloud map.
Within the context of subterranean environments, we expand the applicability of our framework by introducing a modular architecture capable of adapting to specific environments based on their unique features. This adaptability is crucial to tackle challenges posed by subterranean settings, which include reduced lighting, confined spaces, and irregular structural elements.
Furthermore, our framework overcomes the constraints of typical re-localization methods reliant on a single global point cloud map file as input. These methods typically necessitate a manual initial guess or suffer from extended computational times. In contrast, our framework eliminates the need for a manual initial guess and significantly reduces computational time.
Of particular importance, our proposed approach acknowledges the critical role of junctions, a key factor in establishing a complete autonomous pipeline. Junctions play a pivotal role in autonomously triggering the re-localization process, ensuring seamless navigation in complex subterranean environments.
Through these advancements and adaptations, our proposed re-localization framework offers a comprehensive and efficient solution. It combines the strengths of place recognition methods and integrated re-localization solutions while mitigating their respective limitations. This holistic approach contributes to improved accuracy, flexibility, and computational efficiency across various environments, ultimately leading to enhanced re-localization performance.

\subsection{Contributions}

The contributions of this work can be summarized as follows:
\begin{enumerate}
    \item We present a comprehensive framework for re-localization within a given global 3D point cloud map. Our approach capitalizes on data-driven descriptors to facilitate efficient place recognition. Significantly, as depicted in Table~\ref{table:comparison}, our framework integrates the capability to explore the top-k candidates, thereby augmenting the overall system's resiliency.
    \item By conducting comprehensive real-world field experiments, we showcase the strong performance of our framework in achieving reliable re-localization within demanding subterranean environments. Notably, our approach sets itself apart by eliminating the requirement for a manual initial estimate, a feature that distinguishes it from other ROS-based packages utilizing comparable input formats~\cite{ros}, including relocalization methods like LIO-SAM~\cite{LIO-SAM}, FAST-LIO~\cite{FAST-LIO}, and hdl global localization~\cite{hdl_localization}.
    \item We present a novel modular architecture that empowers descriptors to adapt to specific events or tasks. This is accomplished by integrating a classification module, which enables the robot to autonomously initiate the re-localization process through event detection, such as recognizing a junction. This strategy enhances the probability of successful outcomes and extends the robot's exploration autonomy.
    \item We showcase a data-handling process that facilitates the transition from large public datasets and place recognition solutions to learning from a limited amount of field mission data. Additionally, we propose a direct semantic global re-localization approach, which, to the best of the authors' knowledge, is absent from the current literature. We describe various techniques employed in this process, including joint training, negative mining, fine-tuning and label smoothing.
\end{enumerate}
Overall, these contributions provide a comprehensive and innovative framework for re-localization, addressing key challenges in the field and demonstrating improved performance in challenging environments.

\begin{figure*}[t!] 
    \centering
    \includegraphics[width=\textwidth]{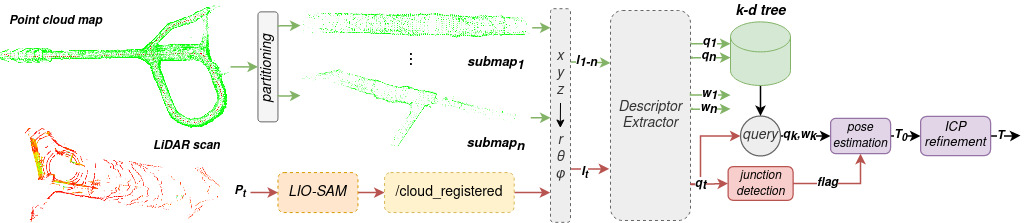}
    \caption{The overall pipeline of the proposed architecture. The red arrows follow the workflow of the current scan $\mathcal{P}_t$, while the green arrows follow the workflow of the given point cloud map.}
    \label{fig:architecture}
\end{figure*}


\section{The proposed approach} \label{sec:approach}

The goal of this article is to introduce a global re-localization algorithm that is able to yield a rigid transform $T \in SE(3)$ so that the current robot frame $\mathcal{R}$ is transformed to the global map frame $\mathcal{M}$. Considering a robot $r$ operating in $\mathbb{R}^3$ space, it generates 3D point cloud LiDAR scans, $\mathcal{P} \in \mathbb{R}^3$, with respect to the robot frame $\mathcal{R}$. Given a known point cloud map $M$ in the global map frame $\mathcal{M}$ and its corresponding trajectory $Tr$, denoted as:
\begin{equation}
\label{eq:map_trajectory}
\begin{split}
    M &= \{m_1, m_2, \ldots, m_n\}, \\
    Tr &= \{\;p_1, \;p_2, \ldots,\;p_k\},
\end{split}
\end{equation}
where $m\in\mathbb{R}^3$ are sets of points and $p\in\mathbb{R}^3$ are sets of poses $p_k = (x_k,y_k,z_k)$, we are looking for the homogeneous rigid transformation of the special Euclidean group, defined as:
\begin{equation} \label{eq:transform}
    T = \left[ \begin{array}{cc}
         R & \vec p\\
         0 & 1
    \end{array} \right] \in SE(3),
\end{equation}
where $R \in SO(3)$ is the rotational matrix and $\vec p \in \mathbb{R}^3$ is the translational vector. 
To tackle this problem and acquire the transform $T$, we follow the steps depicted in Figure~\ref{fig:architecture}, which can be summarized as: a) Map Partitioning, b) Point Cloud Projection, c) Descriptor Extraction, d) Initial Pose Estimation, and e) Pose Refinement.

\subsection{Map Partitioning}

To work with a point cloud map, we must partition it into individual scans. Let the point cloud map be denoted as $M$, and let the trajectory of the robot be represented as a sequence of $n$ points, $\{p_1, p_2, ..., p_n\}$. We create a $k$-d tree database of the visited places to enable efficient search using descriptors. We partition $M$ into $n$ scans by transforming the point clouds according to the corresponding trajectory points. Let $M_i$ denote the point cloud map for the $i^{th}$ scan. We transform $M$ according to $p_i$ to obtain $M_i$, such that each point $m$ in $M_i$ is expressed with respect to the robot frame and not the map frame. Mathematically, we can write:
\begin{equation}
    M_i = \{T_{p_i} (m) \ | \ m \in M\},
\end{equation}
where $T_{p_i}(m)$ denotes the transformation of point $m$ according to $p_i$. Essentially, we are creating a $k$-d tree database of the visited places, which we can later search efficiently with the descriptors. 
By providing the point cloud map along with the discrete trajectory, we partition the map into $n$ scans, where $n$ is the number of points in the trajectory.
Then we transform the point clouds according to the corresponding trajectory points, so we always have the partitioned map points in respect to the robot frame and not the map frame. 

\subsection{Point Cloud Projection}

\begin{figure*}[h!]
    \centering
    \includegraphics[width=\textwidth]{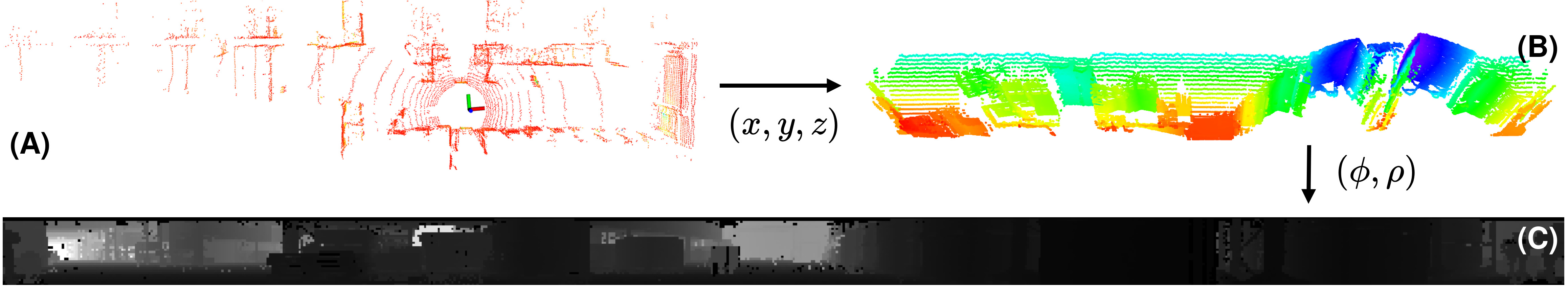}
    \caption{(A) The original LiDAR scan $\mathcal{P}{k}$. (B) The result of the spherical projection. (C) The depth image $\mathcal{I}{k}$ obtained by mapping the range of each point to its corresponding image pixel.}
    \label{fig:projection}
\end{figure*}

The primary purpose of the Point Cloud Projection submodule is to convert the LiDAR point cloud scan data $\mathcal{P}_{k}$ at each time step $k$  or each submap $M_k$, into a 2D depth image $\mathcal{I}_{k}$ using a spherical projection model. This transformation is accomplished by projecting a list of point coordinates $p_x$, $p_y$, and $p_z$ onto a 2D spherical grid, as illustrated in Fig.~\ref{fig:projection}. The pixel value of each point in the grid is determined by its range $\rho$ from the sensor's frame, as described by the equations:
\begin{equation}
\begin{array}{c}
    \phi = atan(\frac{p_x}{p_y}) \\
    \\
    \rho = asin(\frac{p_x^2 + p_y^2}{\sqrt{p_x^2 + p_y^2 + p_z^2}})
\end{array}
\end{equation}
Converting LiDAR point cloud scans into range images has significant advantages over using raw point cloud data. The panoramic $360^o$ view obtained from range images enables the production of orientation-invariant descriptors that can be used for various applications, including object detection and classification. Using a 2D CNN on range images can be particularly beneficial because of their computational efficiency and ability to handle large datasets. Additionally, 2D CNNs are suitable for capturing translational invariance in the data, which is useful for detecting and classifying features from different viewpoints. However, there are some limitations to using range images. One of the main disadvantages is that they capture a less dense view of the surroundings compared to depth sensors, which can make it challenging to extract detailed features from converted range images, especially for tasks that require high accuracy and precision.
\subsection{Descriptor Extraction}

The Descriptor Extraction module is responsible for generating a concise representation of the surrounding topological characteristics, which includes place recognition, orientation regression, and classification information. In this study, we employ a Convolutional Neural Network (CNN) that takes 2D range images, $\mathcal{I}_k$, as input and produces $2 \times 64$ vector sets, namely $\vec q$ and $\vec w$, which will be further discussed in subsection~\ref{subsec:network} and are denoted as:
\begin{align}
    Q &= \, \{\vec q_1, \vec q_2, \ldots, \vec q_n : \vec q_i \in \mathbb{R}^{64}\}, \\
    W &= \,\{\vec w_1, \vec w_2, \ldots, \vec w_n : \vec w_i \in \mathbb{R}^{64}\}
\end{align}
The vector $\vec q$ captures orientation-invariant information that is specific to each place, while the vector $\vec w$ serves as a compact representation of rotation-variant information. The latter is utilized for estimating the yaw discrepancy in a subsequent stage of the pipeline~\cite{OREOS}.
To build a comprehensive understanding of the environment, we utilize the place-specific vectors $Q$, generated from the partitioned map scans $M_i$. 
By employing the aforementioned approach, we extract essential information from the topological characteristics, allowing for effective place recognition, orientation regression, and classification tasks.

\subsection{Initial Pose Estimation}

In the Initial Pose Estimation module, we make use of the aforementioned descriptive vectors to construct a $k$-d tree. Subsequently, with the current scan from the robot $\mathcal{P}_t$, we predict the vector $\vec q_t$ and query the $k$-d tree to identify nearby potential places. The querying process tries to identify the vector $\vec q_i$ from the vector set $Q$ that has the minimum distance from the vector $\vec q_t$, in the vector space. The process is denoted as:
\begin{equation} \label{eq:argmin}
    i = \operatorname*{arg\,min}_{i \, \in \, \mathbb{N}}f({q_i},q_t) {\text{ where } q_i \in Q},
\end{equation}
where $f(a,b)$ is a function that returns the Euclidean distance between two multidimensional vectors $a$ and $b$, described as:
\begin{equation}
    f(a,b) = \sqrt{(a_1 - b_1)^2 + (a_2 - b_2)^2 + \cdots + (a_n - b_n)^2}
\end{equation}
Once the $k$-d tree has been queried with the current vector $\vec{q_t}$, we obtain the indexes of the nearest neighbors, along with the corresponding vectors $\vec{q_i}$ and $\vec{w_i}$ from the vector set $Q$. This retrieval allows us to obtain the trajectory points associated with these neighbors, facilitating the acquisition of the initial translation vector $\vec{p_0}$. The top-$k$ indexes of the nearest neighbors are returned from the search, and the index corresponding to the minimum distance is selected as the primary candidate for re-localization. This choice is based on the assumption that the nearest neighbor with the least distance is more likely to yield an accurate re-localization result. However, it's important to note that in cases where the first candidate fails to provide the desired outcome, the remaining top-$k$ indexes can be utilized as alternative candidates, allowing for resiliency in cases of challenging localization scenarios.
The subsequent step involves feeding the orientation estimation module with the rotation variant vectors $\vec w_t$ and $\vec w_i$. This module is responsible for estimating the yaw discrepancy $\delta \theta$ between the query point cloud and the nearest retrieved candidate from the partitioned map. 
With the calculated $\delta \theta$, we can construct the rotation matrix $R_0(\delta \theta)$, which represents the initial estimation. Combining this rotation matrix with the initial translation vector $\vec{p_0}$, we obtain the complete initial estimation denoted as:
\begin{equation} \label{eq:transform}
    T_0 = \left[ \begin{array}{cc}
         R_0 & p_0\\
         0 & 1
    \end{array} \right] \in SE(3)
\end{equation}
This process allows us to align the query point cloud with the nearest candidate from the partitioned map, providing an initial estimation that incorporates both translation and rotation information.

\subsection{Pose Refinement}

Using the obtained initial pose estimation, we can enhance the accuracy of the pose by utilizing it as a prior condition in the registration algorithm, Iterative Closest Point (ICP)~\cite{icp}. This algorithm iteratively aligns the query point cloud with the nearest candidate from the partitioned map, progressively refining the pose estimation.
Additionally, the distance between the two vectors $\vec q_t$ and $\vec q_k$ can be leveraged to establish a distance threshold for the registration method. By incorporating this threshold, we can aid the registration process by discarding potential matches that exceed the threshold. 
By integrating the initial pose estimation and utilizing the distance information between the vectors, we can achieve a more accurate and efficient refinement of the pose estimation through the ICP registration algorithm.

\subsection{Event-based triggering}

In order to enhance the autonomy of our robots during field exploration missions, we utilize the classification module to identify instances when the robot reaches a junction. This classification capability serves as a crucial trigger for the global re-localization process, particularly in scenarios involving multi-robot exploration, where multiple robots need to share the same map~\cite{stathoulopoulos2023frame}.
By providing the current vector $\vec q_t$ as input, the classifier is capable of distinguishing between different configurations, such as a straight tunnel, a junction, or a turn. The event of reaching a junction serves as the trigger point for initiating the global re-localization process. This is due to the fact that junctions tend to exhibit more distinctive features and offer a higher likelihood of success in accurately re-establishing the robot's global position within the environment.


\section{Neural Network} \label{subsec:network}

According to the aforementioned related work, we have decided to work with a network architecture based on 2D range images generated from 3D LiDAR scans, and not with the point clouds directly, since the deep learning advancements in feature extraction from images have always demonstrated a robust result. Therefore, our network architecture and descriptor extraction process is based on~\cite{OREOS} and is adjusted following the principles described in~\cite{very_deep, image_similarity}, as well as our own proposed addition for driving learned descriptors to certain features. 

\subsection{Network Architecture}
\begin{figure*}[t!]
    \centering
    \includegraphics[width=\textwidth]{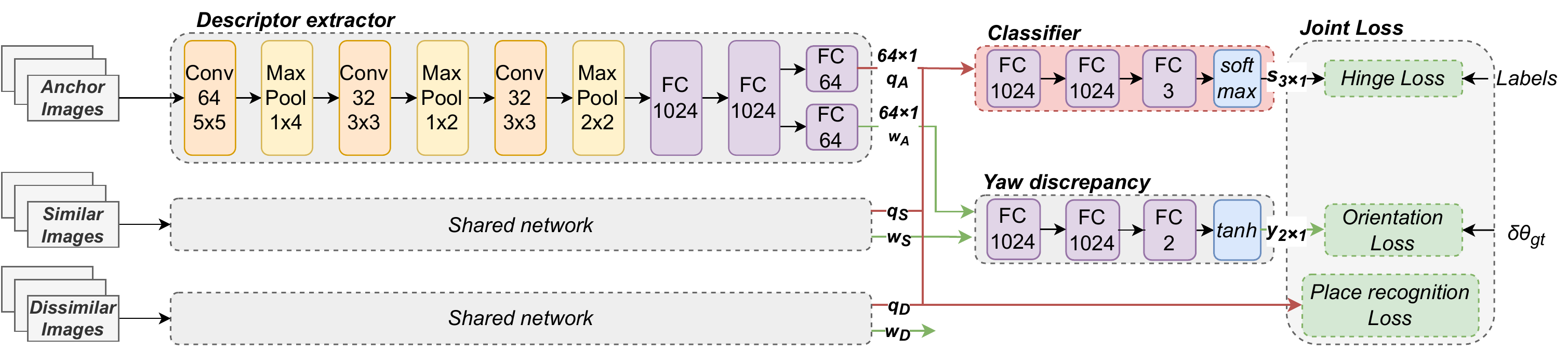}
    \caption{The Siamese network consists of three convolutional layers followed by max pooling and then two fully connected layers, as proposed by~\cite{OREOS}. The number of filters and the kernel size have been adjusted to the considered testing environments and sensor utilized, and in addition we have an extra classifying module that boosts the descriptor extraction process.}
    \label{fig:network}
\end{figure*}

The architecture of the proposed CNN is composed of 2D convolutional layers followed by Max Pooling layers.
Then, the Fully Connected layers compress the features and map them into a compact descriptor representation, as depicted in Figure~\ref{fig:network}.
We start with 64~filters for the first convolutional layer and a 5×5~filter size, since we want a larger area to compensate for the sparse nature of our depth images.
For the other two layers, we move down to 32~filters with a 3×3~filter size.
The output of the \textit{descriptor extractor} are 2×64×1 vectors, $\vec q$ and $\vec w$ respectively. 
As mentioned before, the vector $\vec q$ encodes place dependent information and is invariant to orientation changes, while $\vec w$ is an orientation specific vector and is used to decrease the angle discrepancy~\cite{OREOS}. 
This process is handled by an extra \textit{orientation estimation} module, which takes as an input two vectors $\vec w$ and outputs a 2×1 vector $y_{yaw}$, after two fully connected layers and a $\tanh$ activation function. 
The third module of the presented network is a classifier and is consisted of two fully connected layers and a \textit{softmax} activation function. 
In this case, we use the \textit{classifier} to detect the topological characteristics of the surrounding environment and more specific, to classify among: (a) a straight tunnel, (b) a junction, or (c) a turn. 
The classification process is performed based on a derived vector $\vec q$ and depending on the mission, the \textit{classifier} can be trained on detecting other characteristics, for example pipelines or shafts.

\subsection{Loss functions}

The overall network is composed of three different modules, where each one of them pursues a different goal.
The \textit{descriptor extraction} module needs to find two orthogonal vectors $\vec q$ and $\vec w$. These orthogonal vectors are crucial for achieving distinct objectives within the subsequent modules of the system. The \textit{orientation estimation} module estimates the yaw difference from two compact vectors, $\vec w$, ensuring that this vector's orientation-specific information is separated from other components. Similarly, the \textit{classifier} has the goal of predicting the correct class based on a descriptor vector $\vec q$. This concept of vector orthogonality, as introduced in the work by~\cite{OREOS}, emerges due to the divergent objectives pursued by the two vectors, each striving for distinct and independent goals. Specifically, one vector is designed to be place-dependent while remaining orientation-invariant, whereas the other vector is orientation-specific yet place-independent. The ensuing orthogonal relationship is established as a consequence of these differing design principles. For each of these three goals, a loss function is defined, denoted as $L_{pr}$ for the place-recognition loss, $L_{\theta}$ for the orientation loss, and $L_c$ for the classifier's loss. The choice of orthogonal vectors aligns well with the principle that orthogonal vectors often arise when capturing different aspects or features that are independent of each other. In our case, $\vec q$ and $\vec w$ represent distinct sets of features, each encoding unique characteristics of the data. This design choice ensures that the vectors do not overlap in their information content, preventing redundancy and correlation. By making sure that these vectors are orthogonal, we enhance the ability of your system to focus on the specific aspects relevant to each module's task, thus contributing to the overall efficiency and effectiveness of your approach.

Starting with $L_{pr}$, in order to train our network for the task of place recognition, we use the triplet loss method~\cite{triplet_loss}.
As demonstrated in Figure \ref{fig:network}, we feed the neural network with three types of images: anchor images, which serve as the reference points for comparison, similar images to the anchor images, and dissimilar images, denoted as $I_A, I_S$ and $I_D$ respectively. These anchor images provide a fixed foundation for evaluating the similarities and differences between the other image pairs, playing a crucial role in our approach for descriptor extraction and subsequent analysis.
We also define $d_S$ as the Euclidean distances between the descriptors $\vec q_A$ from $I_A$ and the descriptors $\vec q_S$ from $I_S$. 
The same applies for $d_D$ and the descriptors $\vec q_D$ from $I_D$. 
The loss function is designed so that similar and dissimilar point cloud pairs are pushed close together and far apart in the derived vector space. 
The parameter $m$ is a margin distance for distinguishing between similar and dissimilar pairs. 
The triplet loss is defined as follows:
\begin{equation}
    L_{pr}(d_S,d_D) = \underbrace{||f(I_A)-f(I_S)||^2}_{d_S} - \underbrace{||f(I_A)-f(I_D)||^2}_{d_D} + m 
\end{equation}
The orientation estimation loss $L_{\theta}$ is a regression loss function that is computed based on the \textit{orientation estimation} module's output $y_{yaw}$ and the ground truth $\delta \theta_{gt}$. For predicting the orientation discrepancy, we only make use of the orientation dependent vectors $\vec w_A$ and $\vec w_S$. The orientation loss is defined as:
\begin{equation}
\begin{split}
    L_{\theta}(y_{yaw}, \delta \theta_{gt}) = \frac{1}{2}((y_{yaw,0} &- \cos{(\delta \theta_{gt})})^2 \\
    &+ (y_{yaw,1} - \sin{(\delta \theta_{gt})})^2)
\end{split}
\end{equation}
As mentioned in~\cite{OREOS}, by transforming the ground truth yaw angle $\delta \theta_{gt}$ into the Euclidean space, we avoid the ambiguity between $0-360^{\circ}$ which could lead to false corrections during training.
The last loss function we need to define is the classification loss $L_c$. 
For this, we choose the Hinge Loss function, defined as follows.
\begin{equation}
    L_c(s_j, s_{l_i}) = \sum_{j\neq s_i}\max{(0, s_j - s_{l_i} + 1)}
\end{equation}
Essentially, the Hinge Loss function is summing across all the incorrect classes ($i\neq j$) and comparing the output of the predicted vector $s$ returned for the $j$-th class label (the incorrect class) and the $l_i$-th class (the correct class).

\subsection{Training the descriptors}

In the training process of our network, we employ a technique known as joint training. Joint training involves simultaneously optimizing the weights and parameters of multiple interconnected neural network modules. This approach is beneficial in machine learning training for several reasons. Firstly, joint training facilitates the creation of synergistic relationships between different modules, enabling them to collectively learn and adapt to complex patterns and interactions present in the data. By training the entire network holistically, information and insights obtained from one module can be shared and utilized by other modules, leading to improved overall performance.
Moreover, joint training promotes the development of a cohesive and unified representation of the input data. As different modules influence each other's learning processes, the network can more effectively capture intricate relationships and dependencies within the data, resulting in a more comprehensive understanding of the underlying patterns.
Additionally, joint training can help mitigate issues related to overfitting, as the modules are optimized together, they collectively strive for a balanced solution that generalizes well to unseen data, reducing the risk of overfitting to individual modules. This approach also encourages the network to learn more discriminative and transferable features, enhancing its ability to handle various scenarios and data variations.
In the context of our specific network, joint training enables us to achieve multiple objectives simultaneously, namely, enhancing localization recall, achieving precise yaw angle estimation, and ensuring robust classification performance. By jointly optimizing the weights of the three Neural Networks, we create a unified learning framework that leverages the strengths of each module while fostering synergies between them. This holistic optimization approach contributes to the effectiveness and efficiency of our network's performance across various tasks.
For this, we combine all three loss functions, defined as $L$:
\begin{equation}
    L = L_{pr} + L_\theta + L_c
\end{equation}
We sample the training point cloud data and then based on the margin $m$ and their ground truth poses, we characterize a similar and a dissimilar to the anchor point cloud, in order to prepare the triplets for the three-tuple shared network. 
As a data augmentation step, we randomly rotate the point clouds around the yaw axis, making sure that the orientation between anchor and the similar point clouds is different while still being from a similar place.
The three point clouds are converted to the range images and then are fed to the \textit{descriptor extractor} network that outputs the three corresponding pair-vectors, $(\vec q_A, \vec w_A), (\vec q_S, \vec w_S)$ and $(\vec q_D, \vec w_D)$. 
The three place dependent vectors are used to compute the $L_{pr}$ loss, while $\vec w_A$ and $\vec w_S$ are passed to the \textit{orientation estimation} network and the corresponding output $y_{yaw}$ along with $\delta \theta_{gt}$ are used to compute the $L_\theta$ loss. 
The vector $\vec q_A$ is also passed to the \textit{classifier}, where the output $s$ with the labels are used to compute the $L_c$ loss. 
The combined loss $L$ is then evaluated and with the ADAM~\cite{adam} learning optimizer the weights are updated.


\section{Experiments and Results}
\label{sec:result}

In this section, we will go through the experimental results, starting from evaluating the performance of the Neural Network architecture, as well as comparing it to its base version.
Then we will further evaluate the complete proposed framework and compare it to the existing ROS available solutions. 
All experiments are carried out in real-world settings with a focus on subterranean environments.

\subsection{Neural Network evaluation}
\label{subsec:net_results}

\subsubsection{Datasets}
\begin{figure*}[t!]
    \begin{subfigure}{0.34\textwidth}
    \includegraphics[width=\columnwidth]{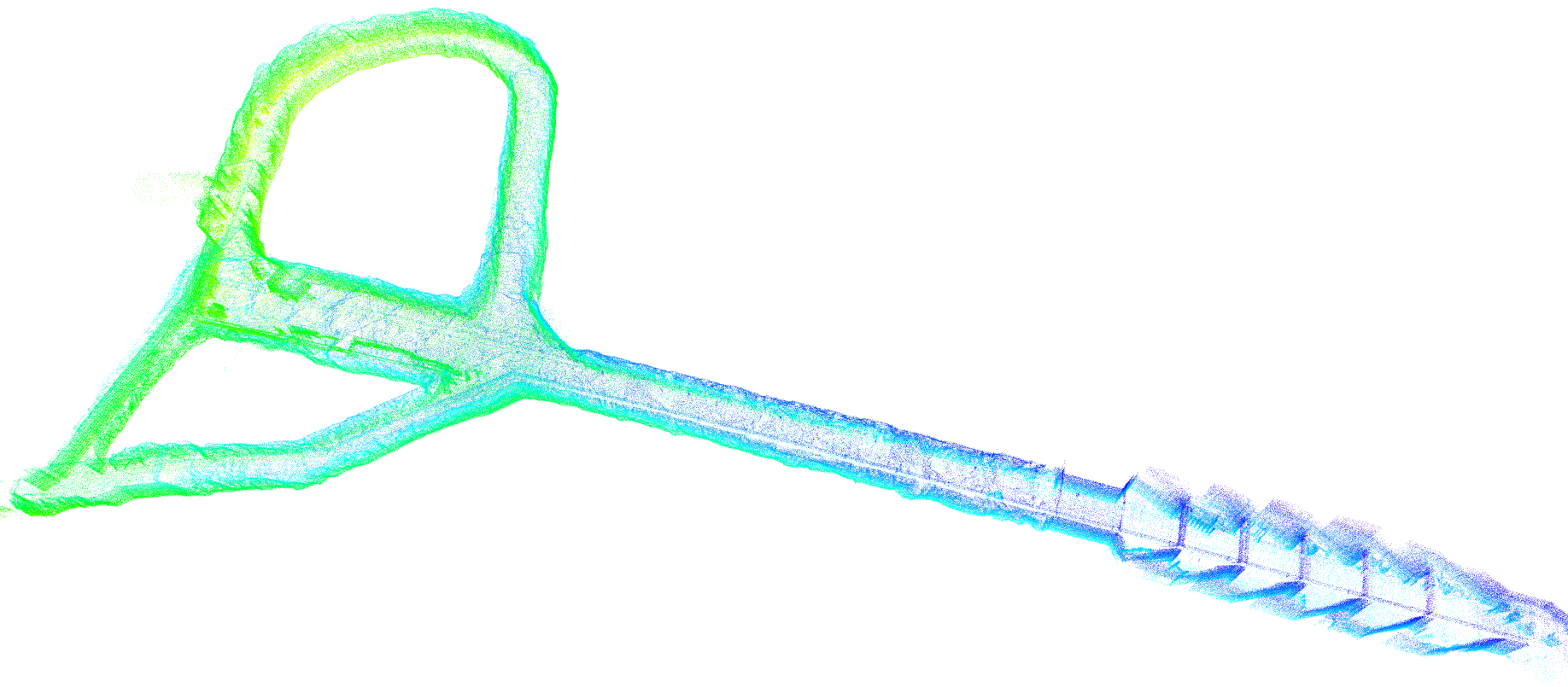}
    \caption*{Map A.}
    \end{subfigure}
    \hspace*{\fill}
    \begin{subfigure}{0.27\textwidth}
    \includegraphics[width=\columnwidth]{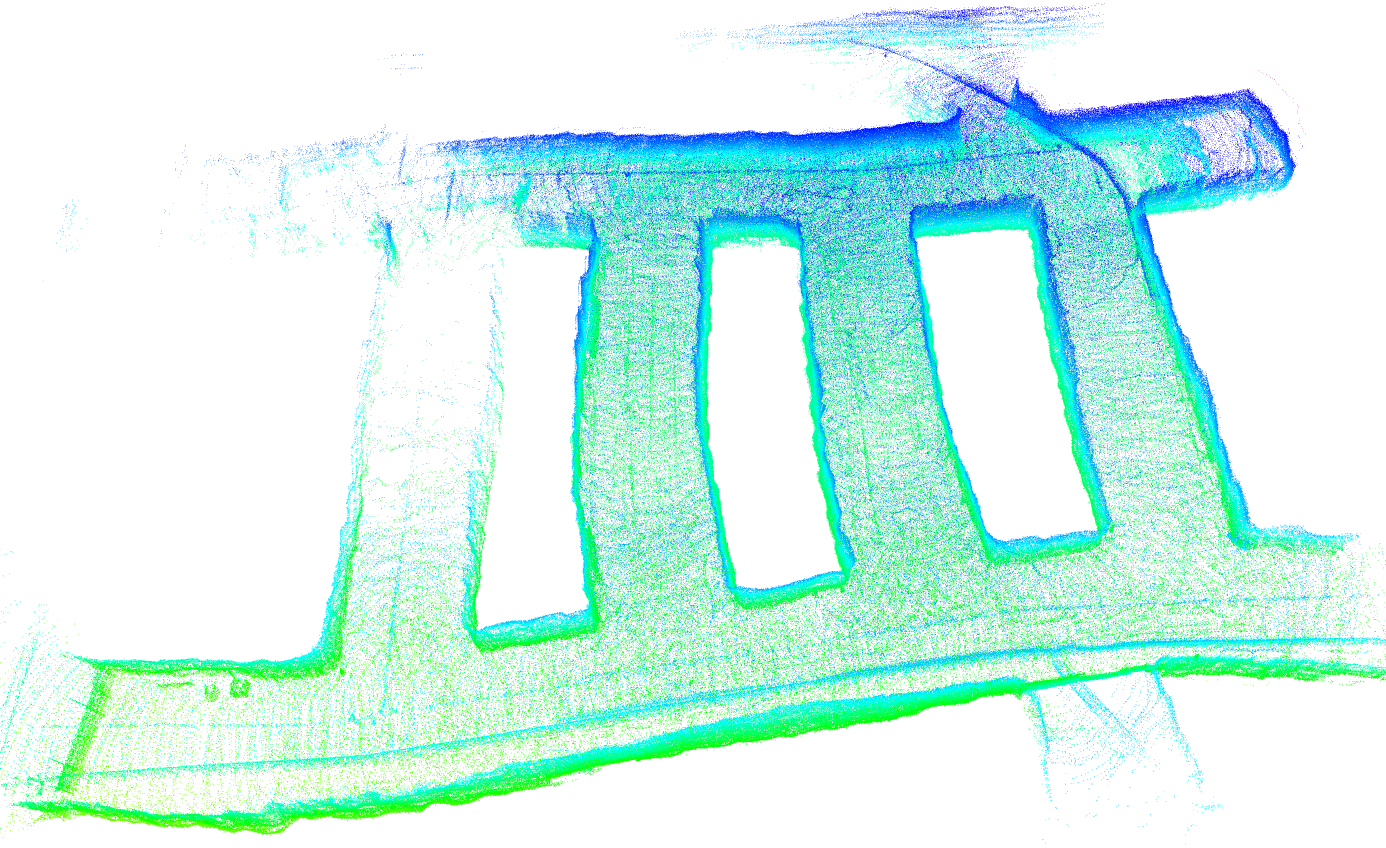}
    \caption*{Map B.}
    \end{subfigure}
    \hspace*{\fill}
    \begin{subfigure}{0.34\textwidth}
    \includegraphics[width=\columnwidth]{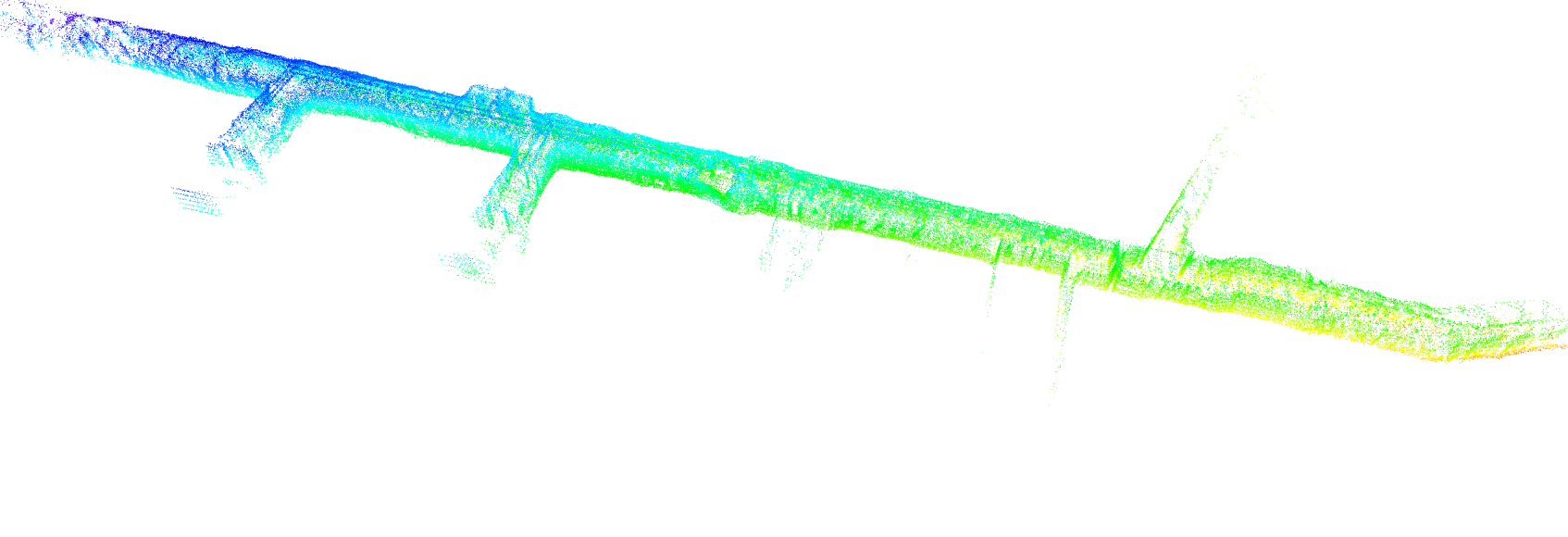}
    \caption*{Map C.}
    \end{subfigure}
    \caption{From map A, only the recordings from the upper branch are used for training, while the rest are used for the evaluation. Map B and C are only utilized for evaluation.}
    \label{fig:pcds}
\end{figure*}
For the training and evaluation process, we use three dataset collections.
The first dataset collection~\cite{KOVAL2022104168} contains recordings from an underground tunnel located in Lule\r{a}, Sweden, as seen in Figure~\ref{fig:pcds}.
For this area, we have recordings from two different robotic platforms. The first is with Spot from Boston Dynamics~\cite{spot}, equipped with an autonomy package~\cite{koval2022experimental}, that includes the Velodyne VLP16 PuckLite 3D LiDAR, the Vectornav VN-100 IMU sensor and an Intel NUC on-board computer. The NUC has an Intel Core i5-10210U, 4-core processor with Intel UHD Graphics and 8 GB of RAM.
The second robotic platform is a custom-built quadrotor~\cite{lindqvist2021compra} and is equipped with the same 3D LiDAR and on-board computer as Spot.
It is important to mention that even though both platforms have similar sensors, the acquired data may differ due to movement noise, dust and accuracy of the IMU, as one is a ground quadruped robot and the other is a flying robot. 
The main difference can be seen on the registered point clouds, as well as the generated range images, since they are operating in different heights and with different form of vibration due to walking or flying.
The cumulative distance covered by the robots within the mapped environment spans approximately 0.5 kilometers, signifying a considerable exploration effort. This extensive traversal yielded a comprehensive representation composed of an approximate total of $3\cdot 10^5$ points, capturing the features and layout of the environment.
The subsequent dataset, represented as Map B in Figure~\ref{fig:pcds}, stems from an authentic subterranean mining site. In contrast to the initial dataset, this collection encompasses more expansive tunnels, reaching widths of up to 10 meters. Within this environment, multiple intricate junctions coexist amidst a relatively featureless setting, presenting a distinct set of challenges for robotic exploration and mapping.
Spanning an area of 0.5 kilometers, this environment contributes to the creation of a map containing an estimated $1.6 \cdot 10^6$ points. The intricate complexities of the larger tunnels and junctions further augment the richness of the mapped data. This dataset serves as a valuable resource for evaluating the efficacy and robustness of our proposed approach within diverse subterranean settings, highlighting its adaptability to a spectrum of real-world environments.
The third and last dataset collection is from the same underground tunnel as the first one but from a different passage, and it is depicted on Figure~\ref{fig:pcds} as Map C.
Within this particular context, the environment takes on a more constrained configuration, resembling a corridor that is both narrower and more linear in nature. This corridor stretches across a length of approximately 150 meters, characterized by its distinct geometric attributes and confined spatial characteristics. In terms of data density, the environment contributes a collection of data points totaling approximately $1.2 \cdot 10^5$. The intricacies of this corridor-like setting pose unique challenges for mapping and navigation, allowing us to examine the adaptability and effectiveness of our proposed approach in constrained and specialized subterranean scenarios.
From all datasets, we make use of the 3D LiDAR scans and the odometry data in order to train our models. 
The labels for training the junction detection module were handcrafted on all datasets. 
It's important to highlight the process behind acquiring the maps A, B, and C showcased in Figure~\ref{fig:pcds}, as these maps lay the foundation for the subsequent re-localization process in our experiments. In both scenarios, the generation of these maps relied on the utilization of LIO-SAM~\cite{LIO-SAM}, a Simultaneous Localization and Mapping (SLAM) algorithm. SLAM is a fundamental capability that empowers autonomous systems to concurrently estimate their own position (localization) and construct a map of the environment they are navigating.
LIO-SAM leverages both 3D LiDAR scans and IMU measurements to perform this dual task. In the case of Map A, manual control was employed to guide the robot and build the map, ensuring a meticulously crafted initial representation. Map B, on the other hand, is a product of merging multiple maps generated from distinct autonomous missions. This map merging process was facilitated by FRAME~\cite{stathoulopoulos2023frame}, a map-merging algorithm. Lastly, the creation of the final map C was orchestrated through the autonomy framework COMPRA~\cite{lindqvist2021compra}. COMPRA facilitated the autonomous exploration of the tunnel, enabling the robot to venture forth, comprehensively map the environment, and safely return.
In essence, SLAM plays a pivotal role in generating accurate maps that serve as the bedrock for subsequent localization and navigation tasks. It allows to concurrently estimate the robot's location and construct a comprehensive map of its surroundings. The combined efforts of SLAM algorithms and autonomous frameworks contribute significantly to the reliable and informed navigation of robotic systems within complex environments.

\subsubsection{Data sampling and training process}

As mentioned in Section \ref{subsec:network}, the neural network is based on the triplet network architecture and therefore requires sampling three tuples of anchor, similar and dissimilar pairs. We consider two point clouds as similar, if their ground-truth poses, defined as $p$, are within $3$~\unit{meters}, $|p_A - p_S| \leq 3$.
In addressing the dissimilar pairs, our approach employs a strategy known as hard-negative mining, to enhance the performance of our network. This technique focuses on selecting dissimilar pairs, which consist of point clouds that are not related to each other. To achieve this, we employ a two-stage negative mining strategy. In the initial stage, we randomly sample point clouds that are beyond a 3-\unit{meter} radius from each other ($|p_A - p_D| \geq 3$), ensuring a significant level of dissimilarity between the selected pairs. As the training process advances, we move on to the later stage, where we sample point clouds within a radius of 3 to 6 meters from the anchor ($3 \leq |p_A - p_D| \leq 6$).
The essence of this strategy lies in the concept of "hard-negative mining," a concept introduced by~\cite{negative_mining}. By introducing progressively more challenging negative samples during the later stages of training, the network is exposed to point cloud pairs that are difficult to distinguish, pushing its boundaries and honing its ability to discern subtle differences. This process of incrementally introducing harder-to-distinguish triplets helps the network adapt and improve its performance in the advanced phases of convergence.
The benefits of negative mining are twofold. Firstly, it exposes the network to a wider range of training examples, helping it learn from diverse scenarios and enhancing its generalization capabilities. Secondly, by focusing on challenging triplets that are initially challenging to differentiate, the network becomes more robust and capable of handling complex and intricate variations in the input data. This strategic integration of negative mining contributes significantly to the network's ability to achieve higher performance levels and better representation learning.
Given the constrained availability of datasets from subterranean environments, we adopt a strategic approach by utilizing data solely from the initial dataset collection for neural network training. Achieving a balanced dataset is crucial, particularly during the training process of the \textit{classifier} module. However, the inherent nature of the dataset introduces a bias toward straight corridors due to the environmental conditions. To address this issue, we adopt the concept of fine-tuning, which proves beneficial for enhancing the network's ability to generalize across diverse scenarios.
To commence this fine-tuning process, we initiate training without the inclusion of the classifying module. This initial phase allows the network to learn essential features and patterns inherent in the subterranean environments, undisturbed by the classification task. Following this, we strategically resample the dataset to achieve balance among the three distinct classes: a) straight corridors, b) junctions, and c) turns. This balanced dataset enables the network to learn from a diverse set of scenarios, capturing the subtleties associated with each class.
Subsequently, we initiate the fine-tuning process, this time enabling the classifying module. By fine-tuning with the balanced dataset, the network refines its understanding of different classes, effectively adapting its learned features to align with the intricacies of subterranean corridors, junctions, and turns. This fine-tuning approach ensures that the \textit{classifier} becomes adept at identifying and classifying different environmental configurations, thereby enhancing the overall accuracy and robustness of our network's performance in real-world scenarios with limited data availability.
Furthermore, given the inherent challenge of accurately determining the initiation and termination points of junctions, we employ a regularization technique known as label smoothing~\cite{goodfellow, label_smoothing} to refine our training process. Label smoothing entails modifying the target labels used in training to be more softly distributed, rather than using the conventional binary 0 and 1 labels. 
\begin{figure*}[t!]
    \includegraphics[width=0.32\textwidth]{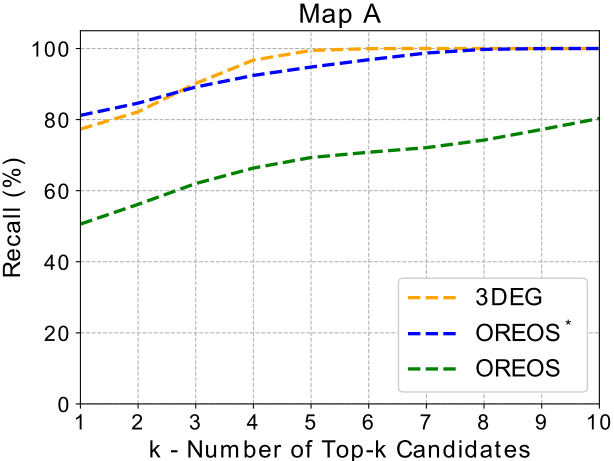}
    \hfill
    \includegraphics[width=0.32\textwidth]{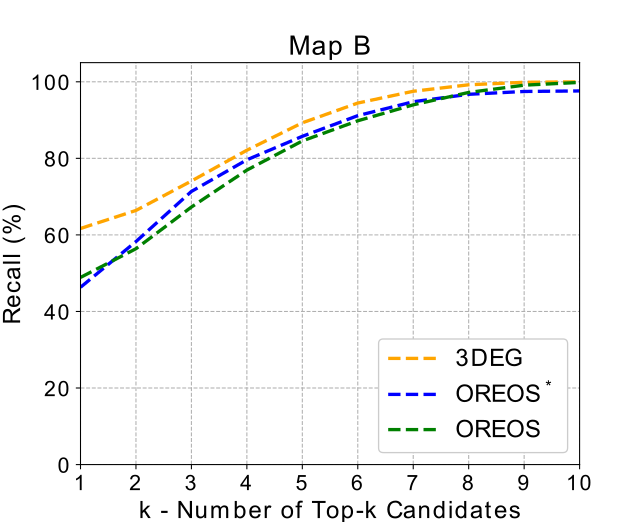}
    \hfill
    \includegraphics[width=0.32\textwidth]{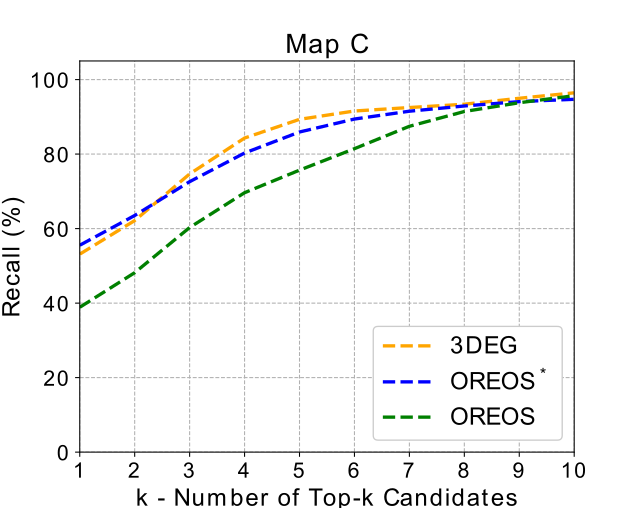}
    \caption{Performance comparison of the proposed framework 3DEG, our implementation of OREOS, as described by Schaupp et al., and our implementation of OREOS but with adjusted network parameters as seen in Figure~\ref{fig:network}, denoted as OREOS$^*$.}
    \label{fig:top_k_results}
\end{figure*}
\begin{equation} \label{eq:label_smoothing}
    labels_{new} = labels_{old} \cdot (1 - a) + \frac{a}{N}
\end{equation}
This technique serves as a powerful tool to address the issues of overfitting and excessive confidence that may arise in the \textit{classifier} module. 
The mechanism of label smoothing, as described by Equation \ref{eq:label_smoothing}, involves transforming the original hard labels ($labels_{old}$) by redistributing a portion ($\frac{a}{N}$) of the label probability mass uniformly among all classes. The parameter $a$ governs the extent of this redistribution, while $N$ represents the number of classes. By introducing this smoothing process, the classifier is encouraged to exhibit more cautious decision-making, mitigating the risk of extreme overconfidence in its predictions.
By embracing soft labels that encapsulate the uncertainty and ambiguity inherent in labeling complex features, we foster a more balanced and adaptive learning process. In essence, when the classifier makes an incorrect prediction, the use of soft labels results in a reduced loss compared to the conventional binary case. Consequently, the model learns from its mistakes in a more gradual manner, leading to a refined understanding of the intricate features and patterns involved. This integration of label smoothing significantly contributes to enhancing the generalization capacity of the model and its performance in scenarios with inherently complex and uncertain decision boundaries.
\begin{table*}[h!]
\caption{Comparison of the mean absolute error and the standard deviation in degrees for the yaw estimation, the recall percentage of the \textit{classifier} and the mean and standard deviation in meters for the final estimated pose.}
\label{table:yaw_results}
\centering
\begin{adjustbox}{width=1\textwidth}
\begin{tabular}{c|ccc|ccc|ccc} \toprule
            & \multicolumn{3}{c|}{Map A}              & \multicolumn{3}{c|}{Map B}                  & \multicolumn{3}{c}{Map C}                \\ \toprule
METHOD      & 3DEG  & OREOS$^*$      & OREOS          & 3DEG           & OREOS$^*$ & OREOS          & 3DEG           & OREOS$^*$      & OREOS  \\
MEAN (deg)  & 14.98 & \textbf{13.53} & 14.47          & \textbf{14.72} & 15.22     & 15.38          & 17.44          & \textbf{16.45} & 16.59  \\
STD (deg)   & 22.34 & 23.14          & \textbf{19.41} & 19.87          & 20.63     & \textbf{19.60} & \textbf{21.49} & 21.90          & 22.18  \\
MEAN (m)    & \textbf{0.23}  & -              & -              & 0.25           & -         & -              & 0.28           & -              & -      \\
STD (m)     & \textbf{0.12}  & -              & -              & 0.17           & -         & -              & 0.19           & -              & -      \\
RECALL (\%) & \textbf{92.4}  & -              & -              & 89.1           & -         & -              & 91.5           & -              & -      \\ \toprule
\end{tabular}
\end{adjustbox}
\setlength{\belowcaptionskip}{10pt}
\end{table*}
\subsubsection{Place recognition results}

An advantage of frameworks like~\cite{OREOS} over the other discussed re-localization frameworks, is that they offer the top-$k$ candidates for the place recognition problem. as seen in Figure~\ref{fig:top_k_results} and in Table~\ref{table:yaw_results}, the localization recall results show that 3DEG outperforms the OREOS in all scenarios, while in some cases the base model performs better than the one with the extra classifying module. The recall percentage is higher on the first map due to being the map that we used part of to train the neural networks. In addition, we notice that for the second map, that contains the most junctions, the junction detection module provides a significant boost on the top-$1$, with more than $10\%$. The results of Table~\ref{table:yaw_results}, for the first two rows, do not include the ICP refinement. It demonstrates the mean error of the yaw discrepancy estimation before the refinement, between the robot’s frame and the chosen submap frame. From our experience, if two point clouds have a high rotational discrepancy (more than 15$^o$-25$^o$), ICP fails to align them properly. On the other hand, after the yaw estimation and the initial pose estimation performed by our framework, the yaw discrepancy will be less than 20$^o$ and therefore the ICP can align them successfully. As expected, there is no major difference in the performance of the yaw estimation, with the mean and standard deviation matching that of OREOS. 

Moreover, we present the mean error of the final estimated pose from the ground truth and the standard deviation. The results for each map arose from running the relocalization process as the robot explores the map, for approximately every meter travelled. In Table~\ref{table:yaw_results}, we only present them for 3DEG since the final estimation is performed by the ICP registration.

\subsection{Global re-localization results}
\label{subsec:reloc_results}

A part of our contribution is that the proposed framework is a complete global re-localization package that works with a given 3D point cloud map and a  trajectory, by utilizing a place recognition framework, and thus we evaluate its performance against the available re-localization ROS packages, mentioned in Section~\ref{sec:related}. In Table~\ref{table:experimental}, we present the time that each package needs to re-localize, as well as the CPU load and the VRAM usage. The BBS engine from the hdl global localization was not able to correctly re-localize in any of the tested places, and consequently was not included in the table. For the FPFH+RANSAC engine, both methods of DIRECT1 and DIRECT7 were tested, and we have included only the fastest one. Even though we noticed higher re-localization times and memory usage than LIO-SAM based re-localization and FAST-LIO localization, it is worth noting that the biggest delay in our pipeline is the final ICP registration for refining the pose, which can be replaced with other faster registration methods like Fast-ICP~\cite{fast_icp} or TEASER++~\cite{Teaser++}. Throughout our experiments, only FAST-LIO localization was able to keep a robust re-localization performance and that only after a very precise initial guess, something that is not required by the proposed 3DEG framework.
\begin{table*}[h!]
\caption{Experimental evaluation of the available re-localization packages, from Map A of Figure~\ref{fig:pcds}, with comparison of the average computational time, CPU load, memory usage and mean error.}
\label{table:experimental}
\centering
\begin{adjustbox}{width=1\textwidth}
\centering
\begin{tabular}{c|cccc|cccc} \toprule
           & \multicolumn{4}{c|}{Starting long corridor}                    & \multicolumn{4}{c}{Lower corridor}                      \\ \toprule
METHOD     & FPFH+RANSAC   & LIO-SAM       & FAST-LIO       & 3DEG          & FPFH+RANSAC   & LIO-SAM       & FAST-LIO       & 3DEG   \\
TIME (sec) & 2.61+22.99    & 0.501         & \textbf{0.229} & 1.331         & -             & -             & \textbf{0.205} & 1.212  \\
CPU (\%)   & 83.5          & \textbf{13.4} & 14.5           & 19.7          & -             & -             & \textbf{9.3}   & 15.3   \\
VRAM (GiB) & \textbf{1.87} & 1.96          & 4.01           & 5.00          & -             & -             & \textbf{4.01}  & 5.00   \\
MEAN (m)   & 0.64          & 0.44          & 0.35           & \textbf{0.30}          & -             & -             & 0.37           & \textbf{0.28}   \\ \toprule
           & \multicolumn{4}{c|}{1\textsuperscript{st} junction}            & \multicolumn{4}{c}{2\textsuperscript{nd} junction}      \\ \toprule
METHOD     & FPFH+RANSAC   & LIO-SAM       & FAST-LIO       & 3DEG          & FPFH+RANSAC   & LIO-SAM       & FAST-LIO       & 3DEG   \\
TIME (sec) & 1.26+24.61    & 1.018         & \textbf{0.164} & 1.294         & 12.62+23.22   & 0.552         & \textbf{0.132} & 1.113  \\
CPU (\%)   & 92.1          & 14.8          & 15.9           & \textbf{13.6} & 86.9          & \textbf{14.2} & 15.5           & 14.5   \\
VRAM (GiB) & \textbf{1.87} & 1.96          & 4.01           & 5.00          & \textbf{1.87} & 1.96          & 4.01           & 5.00   \\
MEAN (m)   & 0.60          & 0.35          & 0.32           & \textbf{0.23}          & 0.62          & 0.39          & 0.36           & \textbf{0.24}   \\ \toprule
\end{tabular}
\end{adjustbox}
\end{table*}

\section{Limitations}
\label{sec:limitations}

Nevertheless, our approach still has limitations. Working in subterranean environments, where the presence of dirt and dust is directly translated into noise, significantly affects the low-resolution VLP16 scans. This results in the degradation of the resolution of the generated range images, making it hard to train the descriptors. Another limitation is the currently used registration method, which can either fail to refine the pose or have a high time and computational cost, especially if the distance threshold is not chosen properly. The angle regression is only present in the yaw angle, providing a 4 DoF initial estimation instead of 6 DoF. To accommodate a different type of environment, re-training is needed with a new classifier, better capturing the features of that environment. Last but not least, as a future step the implementation code should be optimized, which for the moment is not optimal and highly affects the runtime and memory usage of the algorithm.

\section{Conclusions}
\label{sec:conclusion}

In this article, we have introduced the 3DEG framework, which offers a comprehensive solution for global re-localization in a 3D point cloud map setting. This novel framework leverages data-driven descriptors and is designed to autonomously initiate the re-localization process upon detecting a junction within the environment. By incorporating this junction-based triggering mechanism, our framework provides an effective means to address re-localization challenges.
One significant aspect of our proposed framework is its ability to provide resiliency through the inclusion of multiple candidates. This semi-autonomous operation enhances the success rate of critical missions such as search and rescue. By offering multiple candidate options during the re-localization process, our framework increases the robustness and reliability of the system, allowing it to adapt to various scenarios and overcome potential failures.
Overall, the primary objective of this paper is to present a comprehensive re-localization pipeline specifically designed for challenging tunnel environments. Our framework stands out from existing conventional methods, which often rely solely on place recognition or pose estimation approaches that may struggle in such demanding settings. Through our proposed 3DEG framework, we aim to provide an effective and reliable solution for global re-localization in challenging tunnel environments.


\addtolength{\textheight}{-13cm}   

\bibliographystyle{./IEEEtranBST/IEEEtran}
\bibliography{./IEEEtranBST/IEEEabrv,sample}

\end{document}